\documentclass{article} % For LaTeX2e
\usepackage{iclr2025_conference,times}

\usepackage{amsmath,amsfonts,bm}

\def\eqref#1{equation~\ref{#1}}
\def\1{\bm{1}}

\DeclareMathAlphabet{\mathsfit}{\encodingdefault}{\sfdefault}{m}{sl}
\SetMathAlphabet{\mathsfit}{bold}{\encodingdefault}{\sfdefault}{bx}{n}

\usepackage{hyperref}
\usepackage{url}
\usepackage{graphicx}
\usepackage{subfigure}

\title{Evaluation of a Robust Control System in Real-World Cable-Driven Parallel Robots}

\author{Damir Nurtdinov \& Alexei Kornaev\\
Research Center of the Artificial Intelligence Institute\\
Innopolis University\\
Innopolis, Russia\\
\texttt{d.nurtdinov@innopolis.university \& a.kornaev@innopolis.ru} \\
\And
Aliaksei Korshuk \\
Innopolis University \& Coframe\\
Innopolis, Russia \& California, USA\\
\texttt{a.korshuk@innopolis.university} \\
\AND
Alexander Maloletov \\
Innopolis University \& Volgograd State Technical University\\
Innopolis \& Volgograd, Russia \\
\texttt{a.maloletov@innopolis.ru} \\
}

\iclrfinalcopy % Uncomment for camera-ready version, but NOT for submission.
\begin{document}

\maketitle

\begin{abstract}
This study evaluates the performance of classical and modern control methods for real-world Cable-Driven Parallel Robots (CDPRs), focusing on underconstrained systems with limited time discretization. A comparative analysis is conducted between classical PID controllers and modern reinforcement learning algorithms, including Deep Deterministic Policy Gradient (DDPG), Proximal Policy Optimization (PPO), and Trust Region Policy Optimization (TRPO). The results demonstrate that TRPO outperforms other methods, achieving the lowest root mean square (RMS) errors across various trajectories and exhibiting robustness to larger time intervals between control updates. TRPO's ability to balance exploration and exploitation enables stable control in noisy, real-world environments, reducing reliance on high-frequency sensor feedback and computational demands. These findings highlight TRPO's potential as a robust solution for complex robotic control tasks, with implications for dynamic environments and future applications in sensor fusion or hybrid control strategies.
% TODO: create abstract after review
% github: \url{https://github.com/damurka5/RL_CDPR}
\end{abstract}

\section{Introduction}
\label{intro}
Cable-Driven Parallel Robots (CDPR) have unique parameters, which means they can move heavy loads within a fairly large space. Cable suspended robots can be divided into two types: fully constrained configuration, which does not allow a free position and orientation movements of the end-effector; underconstrained configuration, which does not fully restrict the robot. 

This study focuses on a real-world Cable-Driven Parallel Robot (CDPR), an under-constrained physical system, operating under a control system with limited time discretization. 
Our goal is to conduct a comparative analysis of classical and modern control methods and to optimize their time discretization.

\section{Related Work}
\label{related}
\subsection{Classical Approach}
Proportional-Integral-Derivative (PID) controllers are widely used in robotic systems, including CDPRs, due to their simplicity and effectiveness. PID controllers are straightforward to implement and tune, making them accessible for various CDPR applications. In addition, PID controllers do not require a precise mathematical model of the system, which makes them suitable for complex systems such as CDPRs.

\subsection{Reinforcement Learning control approach}
The Actor-Critic Reinforcement Learning (RL) algorithm is a powerful approach to control CDPR. This method combines the strengths of both value-based and policy-based RL techniques.

% Soft Actor-Critic (SAC) is an advanced reinforcement learning algorithm that has been successfully applied to trajectory planning in cable-based parallel robots. SAC is designed to work effectively in continuous action spaces. The SAC algorithm combines the principles of both policy optimization and Q-learning \citep{https://doi.org/10.48550/arxiv.1812.05905}. 
% It introduces a crucial modification to the standard actor-critic framework by incorporating an entropy term in the objective function. This entropy term encourages exploration by incentivizing the policy to maintain a degree of randomness in its actions \citep{https://doi.org/10.48550/arxiv.1812.05905}.

Deep Deterministic Policy Gradient (DDPG) is an off-policy reinforcement learning algorithm designed for continuous action spaces, making it particularly suitable to control CDPR. DDPG combines the strengths of the Deep Q-Networks (DQN) and deterministic policy gradients to handle the complexities of continuous control tasks \citep{Nomanfar2024}.
% The DDPG algorithm employs an actor-critic architecture. The actor network is responsible for determining the best action in a given state, while the critic network evaluates the quality of the action-state pairs. Both networks are implemented using deep neural networks to handle high-dimensional state and action spaces \citep{Nomanfar2024}.
% One of the key advantages of DDPG for CDPR control is its ability to learn continuous control policies directly from high-dimensional sensory inputs. This makes it well-suited for tasks that require fine-grained control over cable tensions to achieve precise positioning and trajectory following in large workspaces.

Trust Region Policy Optimization (TRPO) is an advanced reinforcement learning algorithm designed to address the challenge of stable and efficient policy updates in complex control tasks. The core idea behind TRPO is to maximize the expected return of the policy while constraining the change in the policy at each iteration \citep{schulman2015trust}.

Proximal Policy Optimization (PPO) is a popular reinforcement learning algorithm developed by OpenAI that has shown impressive performance in various control tasks, including those applicable to CDPR. PPO is designed to be simple to implement, sample efficient, and capable of solving a wide range of continuous control problems \cite{https://doi.org/10.48550/arxiv.1707.06347}. 
% PPO builds upon the idea of TRPO, but simplifies the implementation and improves sample efficiency. The core principle of PPO is to update the policy in small steps, ensuring that the new policy does not deviate too far from the old one. This approach helps maintain stability during training while still allowing for significant improvements over time. 

\section{Methodology}
\label{methodology}

\subsection{Control Strategies}
Various control strategies have been explored for CDPRs, including classic PID controllers which tuned with intelligent algorithms \citep{Kelly2005} and Reinforcement Learning (RL) that has emerged as a promising approach for CDPR control and trajectory planning \citep{Bouaouda2024}.

One of the most traditional approaches is the use of classic PID controllers, which have been tuned with intelligent algorithms to enhance performance in dynamic environments. PID controllers are favored for their simplicity and effectiveness in achieving desired positions by continuously adjusting the control inputs based on the error between the desired and actual states. Recent studies have demonstrated the effectiveness of PID controllers in managing cable tensions and ensuring precise movement of the end effector in CDPR applications \citep{7367855}.

In addition to traditional control methods, RL has emerged as a promising approach to CDPR control and trajectory planning. RL algorithms, such as the Deep Deterministic Policy Gradient (DDPG) and Proximal Policy Optimization (PPO), leverage the principles of trial-and-error learning to optimize control policies based on feedback from the environment. These algorithms enable the robot to learn complex behaviors through interactions, making them particularly suitable for dynamic tasks where traditional control methods may struggle \citep{unknown}.

% Although PID controllers offer reliable performance under stable conditions, RL methods can adaptively optimize control strategies in more complex scenarios, such as those that involve dynamic environments. As research continues to advance in this field, the integration of these approaches may lead to more robust and efficient control systems for CDPRs, ultimately improving their applicability in various industrial and research settings.

\subsection{Environment}

The environment for the CDPR is implemented as a custom OpenAI Gym environment, providing a standardized interface for reinforcement learning algorithms. This environment encapsulates the dynamics and control of a 4-cable CDPR system. \footnote{\url{https://github.com/damurka5/RL_CDPR}}.
The observation space is defined as a 12-dimensional continuous space when using target velocity, or a 9-dimensional space without it. It includes the current position, velocity, target position, and optionally, the desired velocity of the end-effector. 
% The boundaries of the observation space are set based on the physical constraints of the CDPR, such as limits of the workspace and maximum speed.

The action space defines control inputs for the four cables, either continuous (normalized forces between -1 and 1) or discrete (specified levels). The step function applies actions, updates the system state using state-space dynamics, and computes rewards based on target distance, proximity, and optionally, velocity. The reset function initializes each episode by randomizing start and target positions, setting velocity to zero, and resetting internal counters.
% The action space represents the control inputs for the four cables. It can be either continuous or discrete, depending on the configuration. In the continuous case, actions are normalized between -1 and 1, representing the force applied to each cable. For the discrete case, actions are discretized into a specified number of levels.

% The step function implements the dynamics of the CDPR. It takes an action as input, applies it to the system, and calculates the new state using a state-space formulation. The function also computes the reward based on several factors, including the improvement in distance to the target, proximity to the target, and optionally, velocity matching.

% The reset function initializes a new episode by randomly sampling the starting position and target position within the workspace boundaries. It sets the initial velocity to zero and constructs the observation state. The function also resets internal variables such as the elapsed steps counter.
% This environment design allows flexible experimentation with various reinforcement learning algorithms, enabling the exploration of different control strategies for the CDPR system.

\section{Kinematics and Dynamics}
\label{Kinematics_and_Dynamics}
Studies have shown that for a CDPR with four cables and non-elastic sagging cables, if the ideal cable model has a single forward kinematic solution, the sagging cable model will also have a single solution \citep{Jean-Pierre-Merlet}. The kinematics equation
for the vector that describes $i^{th}$ cable: %\cite{Bayani}
\begin{align}
\label{eq:kinematics_li}
l_i = c - a_i + R \times b_i
\end{align}
where $c$ is a coordinate of end effector in world coordinate frame, $a_i$ is a cable vanishing point from the $i^{th}$ guide roller, $R$ is a rotation matrix which represents an orientation of the end effector, $b_i$ is a cable connection point. For the first approximation $b_i = \vec{0}$. Jacobian can be calculated as a unit vector $\vec{S_i}$ along $i^{th}$ cable:
\begin{align}
\label{eq:jacobian_1}
J= 
\begin{bmatrix}
	\vec{S_1} &
	\vec{S_2} &
	\vec{S_3} &
        \vec{S_4}
    \end{bmatrix}^T
\end{align}

\section{Results and Discussion}
\label{results}
The following section presents a comparative analysis of the performance of reinforcement learning algorithms DDPG, PPO, and TRPO implemented using the Stable Baselines3 library \footnote{ \url{https://stable-baselines3.readthedocs.io/en/master/}} along with a traditional PD controller, highlighting their effectiveness in controlling cable-driven parallel robots under varying conditions.

We worked with an underconstrained robot configuration that has four cables attached to the servomotors on a fixed $2.31m \times 2.81m$  frame, the anchor points are $3.22 m$ high. Each cable is connected to the box-shaped end effector and a servomotor drum through the pulley.

We created a force control mechanism to run CDPR on different trajectories. We considered the end effector as a point mass of 1 kg, drums and pulleys with zero inertia. After that we have made an environment based on Gymnasium Python classes \footnote{\url{https://gymnasium.farama.org/}}, the reward is calculated from two components: the distance improvement term multiplied by 50 (empirical investigation) and the normalized proximity term multiplied by 5. The training process for all RL algorithms was implemented on a model with $\Delta t=0.1$ sec.

\subsection{DDPG}
The DDPG agent is initialized with a multilayer perceptron (MLP) policy and configured with customizable hyperparameters such as learning rate, buffer size, batch size, and discount factor. A cosine learning rate schedule with warmup is employed to adaptively adjust the learning rate during training. 

The training process for the DDPG algorithm in the CDPR environment demonstrated significant improvements in agent performance over 1.8 million episodes. Initially, the average episode length increased to approximately 28 steps per episode. This growth indicates that the agent was learning to sustain its control actions effectively to achieve better outcomes. However, as training progressed, the episode length converged to an average of 22 steps per episode. 
 
In terms of rewards, the agent's performance improved steadily throughout the training process. By the end of training, the average episode reward reached up to 2000. The increasing reward trend indicates that the agent successfully learned to navigate the complex dynamics of the environment and optimize its control strategy.
% These results highlight the effectiveness of the DDPG algorithm in handling continuous action spaces and learning efficient policies for controlling CDPRs. The convergence of episode lengths and high rewards further demonstrate that the trained agent achieved a balance between exploration and exploitation, enabling it to perform reliably in this challenging robotic control task.
\subsection{PPO}
The PPO agent, as it was in DDPG, is initialized with a multilayer perceptron (MLP) policy and configured with customizable hyperparameters. A cosine learning rate schedule with warm-up was also employed.

The training process for the Proximal Policy Optimization (PPO) algorithm demonstrated significant differences between continuous and discrete action spaces over 7000 episodes. For continuous PPO, the average episode length converged to approximately 22 steps per episode. In contrast, the discrete PPO initially increased to about 21 steps per episode before converging to a lower average of 12 steps.
% This disparity in performance suggests that PPO is more effective when operating with a discrete action space for this particular CDPR control task. The ability of the discrete PPO to complete episodes in fewer steps indicates a more efficient control strategy. This improved performance may be attributed to the algorithm's ability to select actions from a finite set, simplifying the decision-making process and potentially leading to more consistent and optimal choices.
The reward outcomes further underscore the advantage of the discrete action space. While the continuous PPO achieved an impressive episode reward of up to 3000 by the end of training, the discrete PPO significantly outperformed it with rewards reaching approximately 7500. This substantial difference in reward accumulation highlights the discrete PPO's superior ability to optimize the control policy for the CDPR system, resulting in more precise and efficient movements that better satisfy the task objectives.

\subsection{TRPO}

We used the same training process technique for the Trust Region Policy Optimization (TRPO) algorithm, as well as we used the cosine learning rate schedule with warmup. 

The training process for the TRPO algorithm demonstrated impressive performance over 6000 episodes. The average episode length converged to approximately 10 steps per episode. This low number of steps indicates that TRPO quickly learned an efficient control strategy for the CDPR system.
% Rapid convergence to such a low number of steps suggests that TRPO was particularly effective in optimizing the policy for this control task. By maintaining a trust region during policy updates, TRPO likely achieved stable and consistent improvements throughout the training process. This resulted in a policy that could efficiently guide the CDPR to its target position in a minimal number of actions.
The episode reward reached up to 10000 by the end of the training process, which is a significant improvement compared to the results observed for other algorithms like PPO and DDPG. This high reward value indicates that TRPO not only learned to complete the task quickly but also with high precision and efficiency. The combination of low average episode length and high reward suggests that TRPO developed a policy that could accurately control CDPR while minimizing unnecessary movements and optimizing the path to the target position.

After the main training process, we implemented two key changes on pre-trained model for TRPO algorithm: the addition of initial velocity to the state representation and prioritizing initial points closer to the target position.
% By incorporating initial velocity into the state space, the TRPO algorithm gained more comprehensive information about the CDPR's dynamic state, allowing for more informed decision-making from the start of each episode. The prioritization of initial points closer to the target position was introduced to overcome the difficulty the model faced in its early training steps when dealing with widely separated start and end points. This modification helped the algorithm to learn more effectively by gradually increasing the complexity of the control task. 
As a result of these training adjustments, the TRPO model was able to overcome the initial learning hurdles and develop a more robust control policy for the CDPR system.

\begin{figure}[h]
\begin{center}
\subfigure[]{\includegraphics[width=0.49\textwidth]{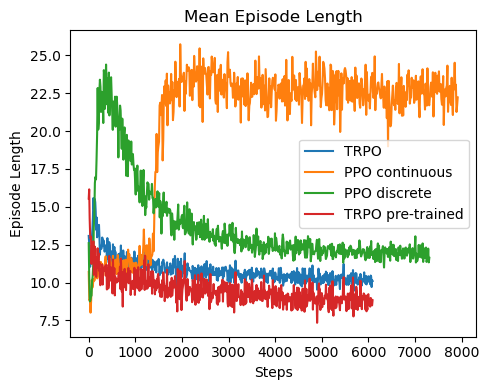}} 
\subfigure[]{\includegraphics[width=0.49\textwidth]{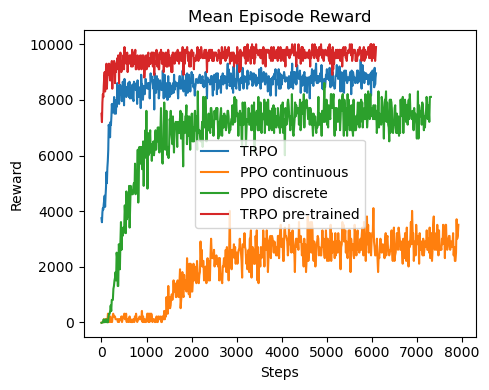}}

\end{center}
\caption{training metrics}
\label{training-metrics}
\end{figure}

The results of the best-performed reinforcement learning algorithms are shown in Fig. \ref{training-metrics}. This figure illustrates performance metrics, such as the mean episode length (a) and reward accumulation (b).
Evaluation of the reinforcement learning algorithms and PID controller on three different trajectories is shown in Table \ref{model-comparison-table}. TRPO consistently achieves the lowest errors, outperforming the other models and the PID controller on all trajectories.
\begin{table}[t]
\caption{RL models and PID controller RMS Errors [m]}
\label{model-comparison-table}
\begin{center}
\begin{tabular}{llllll}
\multicolumn{1}{c}{\bf Trajectory}  &\multicolumn{1}{c}{\bf DDPG} &\multicolumn{1}{c}{\bf PPO} &\multicolumn{1}{c}{\bf TRPO} &\multicolumn{1}{c}{\bf PID}
\\ \hline \\
Circle & 0.0098 & 0.0113 & 0.0075 & 0.0489\\
Spiral 1 & 0.0235 & 0.0125 & 0.0079 & 0.0149 \\
Spiral 2 & 0.0114 & 0.0133 & 0.0084 & 0.0167 \\
\end{tabular}
\end{center}
\end{table}

\subsection{Model simulations on different $\Delta t$}
We have conducted several experiments to check the sustainability and robustness of control strategies on different simulation time intervals, and used optimal gains for PID controller for each experiment. 
\begin{figure}[h]
\begin{center}
\subfigure[]{\includegraphics[width=0.49\textwidth]{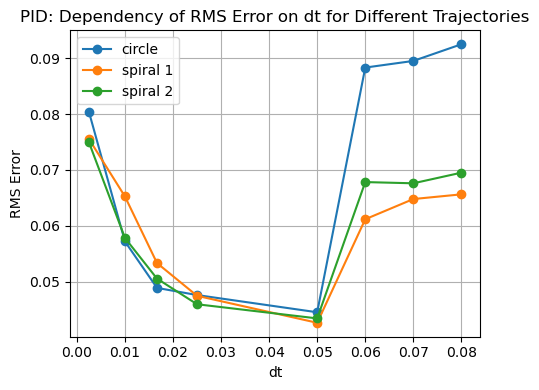}} 
\subfigure[]{\includegraphics[width=0.49\textwidth]{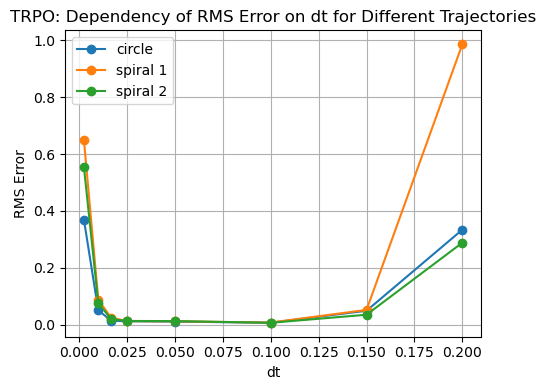}}
% \fbox{\rule[-.5cm]{0cm}{4cm} \rule[-.5cm]{4cm}{0cm}}
\end{center}
\caption{different $\Delta t$ simulations}
\label{dt-experiments}
\end{figure}
The results shown in Fig. \ref{dt-experiments} show that RL algorithm has learned the robot's behavior and can operate at larger $\Delta t$ values than a usual PID controller. 
% \section{Limitations}
% \label{limitations}
% This study presents several limitations that should be considered when interpreting the results. First, we utilized a first approximation model of the CDPR, which simplifies the physics involved in the system. This model does not account for complex dynamics such as cable elasticity, sagging, and other non-linear effects that may significantly influence the robot's performance in real-world scenarios. As a result, the findings may not fully generalize to more sophisticated physical models of the CDPR.

% Additionally, the training process was not tested on a complex physics model, which limits our understanding of how well the algorithms would perform under more realistic conditions. The simplified environment may have led to an overestimation of the algorithms' effectiveness, as they were optimized for a less challenging task. Future work should involve testing these reinforcement learning approaches on a more detailed physics model to better assess their robustness and adaptability in practical applications. Addressing these limitations will be crucial for advancing the deployment of reinforcement learning techniques in real-world CDPR systems.
\section{Conclusion}
\label{conclusion}
The TRPO algorithm demonstrated superior performance compared to DDPG and PPO in both continuous and discrete settings, showcasing its potential for complex robotic control tasks. Its stability with larger time intervals ($\Delta t$) makes it particularly suitable for real-world cable-driven parallel robots, where sensor noise and latency limit high-frequency precision. Unlike PID controllers, TRPO reduces computational demands and adapts to real-world imperfections, making it a strong candidate for dynamic environments. Future work will include physical experiments on real robots to further validate these findings and enhance their practical applicability.
% The TRPO algorithm has shown the best performance compared to other algorithms evaluated in this study, including DDPG and PPO in both continuous and discrete versions. In general, the findings highlight the potential of TRPO as a robust solution for complex robotic control tasks, paving the way for further research and application in real-world scenarios.

% TRPO's ability to maintain stable control with larger time intervals ($\Delta t$) makes it well-suited for real-world cable-driven parallel robots, where sensor noise or latency limits high-frequency precision. Unlike PID controllers, TRPO reduces computational demands and adapts to real-world imperfections, positioning it as a strong candidate for dynamic environments.
\section{Acknowledgement}
All authors were supported by the Research Center of the Artificial Intelligence Institute of Innopolis University.

\bibliography{iclr2025_conference}
\bibliographystyle{iclr2025_conference}

% \appendix
% \section{Appendix}

\end{document}